\documentclass{article}

\usepackage{microtype}
\usepackage{graphicx}
\usepackage{subcaption}
\usepackage{booktabs} 

\usepackage{hyperref}

\newcommand{\best}[1]{\textbf{#1}}
\newcommand{\second}[1]{\underline{#1}}

\usepackage[accepted]{icml2026}

\usepackage{amsmath}
\usepackage{amssymb}
\usepackage{mathtools}
\usepackage{amsthm}
\usepackage{multirow}
\usepackage{makecell}
\usepackage{dsfont}
\usepackage{threeparttable}

\usepackage[capitalize,noabbrev]{cleveref}

\theoremstyle{plain}

\theoremstyle{definition}

\theoremstyle{remark}

\usepackage[textsize=tiny]{todonotes}

\icmltitlerunning{Are We Overconfident in Models and Results for Semi-Supervised 3D Medical Image Segmentation?}

\begin{document}

\twocolumn[
  \icmltitle{Are We Overconfident in Models and Results for Semi-Supervised 3D \\ Medical Image Segmentation?}



  \icmlsetsymbol{equal}{*}

  \begin{icmlauthorlist}
    \icmlauthor{Jun Li}{swjtu}
    \icmlauthor{Ziwei Qin}{swjtu}
  \end{icmlauthorlist}

  \icmlaffiliation{swjtu}{Institute of Systems Science and Technology, School of Electrical Engineering, Southwest Jiaotong University, China}

  \icmlcorrespondingauthor{Jun Li}{dirk.li@outlook.com}

  \icmlkeywords{semi-supervised Learning, Reliability Estimation, Medical Image Segmentation, ICML}

  \vskip 0.3in
]



\printAffiliationsAndNotice{}  

\begin{abstract}
    Semi-supervised learning has become a dominant paradigm for reducing annotation costs. However, we argue that the current progress is clouded by a twofold overconfidence problem.
    Algorithmically, mainstream pseudo-labeling frameworks often conflate prediction confidence with uncertainty, leading to severe confirmation bias.
    Strategically, since multiple benchmark datasets lack dedicated validation sets, some studies use the test set for validation as well, leading to inflated performance estimates. 
    Subsequent methods, compelled to employ the same strategy to surpass reported SOTA, trigger an arms race of overfitting. 
    This raises concerns that the impressive numerical gains in the community may reflect overfitting rather than genuine progress.
    Thus, we propose a tri-space calibrated segmentation framework founded on a principled dual-axis reliability assessment engine. 
    It explicitly decouples confidence from uncertainty and uses this signal to detect and correct confirmation bias across feature, probability, and image spaces in a collaborative manner. 
    Across three benchmark datasets, TCSeg consistently delivers strong performance under existing evaluation protocols. 
    More importantly, we advocate that the community report final-checkpoint results under multiple-run protocols, thereby establishing more rigorous benchmarks with a more realistic perspective.
    Code will be available: \url{github.com/DirkLiii/TCSeg}.
\end{abstract}

\section{Introduction}
\begin{figure}[t]
\centering
\includegraphics[width=0.98\columnwidth]{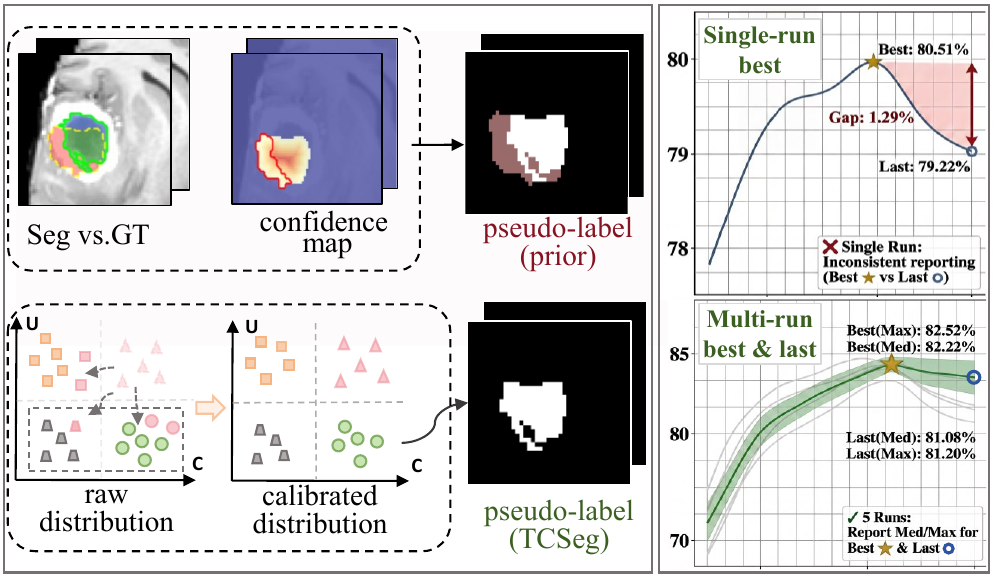}
\caption{
Illustration of overconfidence. Left: miscalibrated pseudo labels induce confirmation bias. Right: single-run best-checkpoint reporting can overstate progress.
}
\label{fig:fig1}
\end{figure}

Volumetric segmentation is pivotal for quantitative medical image analysis, underpinning tasks ranging from surgical planning to postoperative assessment \cite{10982257,wang2021annotation}. While supervised deep learning has achieved parity with human experts \cite{long2015fully}, its deployment is severely bottlenecked by the scarcity of annotations. Given the cost and expertise required for labeling, semi-supervised learning (SSL) has become a common route by exploiting large collections of unlabeled scans alongside sparse supervision \cite{luo2021efficient,yu2019uncertainty}.

Most modern SSL pipelines adopt a self-training framework, using consistency regularization or pseudo-labeling to get supervision from unlabeled samples \cite{9761681,9757875,qiu2023federated,luo2021semi,yu2019uncertainty,qiao2022semi}. These methods often hinge on a fundamental, yet often unverified, shortcut: model’s predictive confidence is a reliable proxy for the correctness of its pseudo-labels. However, deep networks are often poorly calibrated \cite{guo2017calibration} and can assign high probabilities to incorrect predictions. When confidence is used as a gatekeeper for pseudo supervision, miscalibration can become a structural failure mode: errors made early are selected, reinforced, and gradually entrenched---a typical form of confirmation bias.

A common remedy is to inject uncertainty estimation into SSL, e.g., via Monte Carlo dropout or predictive entropy. Yet many such strategies implicitly compress uncertainty into a single scalar derived from the predictive distribution, and thereby blur a crucial distinction. Confidence is a point estimate of preference, uncertainty reflects the strength and consistency of evidence supporting that preference. A voxel can be high-confidence yet high-uncertainty—a scenario typical of confidently wrong predictions caused by overfitting to spurious features \cite{10177922}. By failing to distinguish between these states, many methods inadvertently sanction unreliable pseudo-labels, confusing low-entropy predictions with genuine epistemic reliability.

As shown in Fig.~\ref{fig:fig1}, we first argue that overconfidence is rooted in the conflation of confidence and uncertainty. The softmax score mainly provides a single-model confidence estimate and is often sensitive to data-dependent ambiguity, such as noisy or ambiguous boundaries. However, it does not reliably capture the model’s lack of knowledge. In contrast, epistemic uncertainty is more naturally reflected by disagreement across views, branches, or representations. Thus, we adopt a dual-axis reliability view with two complementary signals: (i) confidence (the consensus of mean predictions) and (ii) uncertainty (the divergence of cross-view evidence). This decoupling allows us to identify and isolate the most damaging subset of pseudo-labels: those that are deceptively confident but structurally unstable. Besides, we highlight a second, community-level source of optimism. As several widely used benchmark datasets lack dedicated validation sets, some studies use test sets instead, which can overstate performance. As such numbers become the reference for SOTA, later work is pushed toward the same protocol to stay comparable, turning evaluation into a test-overfitting race rather than a measure of real progress.

Thus, we propose the Tri-space Calibrated Segmentation (TCSeg) framework, which is organized around a shared reliability engine that explicitly decouples confidence from uncertainty and uses this signal to correct confirmation bias across spaces. Unlike heuristic add-ons, TCSeg orchestrates a collaborative correction mechanism. In probability space, we enforce confidence-band regularization to bound the variance of multi-branch predictions. In feature space, we project voxels onto class-wise prototypes, filtering out predictions that are statistically confident but semantically disjoint from the class manifold. In image space, we deploy a reliability-driven perturbation module that targets cognitive blind spots, forcing the model to learn features in regions identified as high-uncertainty. Finally, to reduce evaluation optimism, we report median and extrema across multiple runs for both best and final checkpoints. This protocol helps separate genuine gains from lucky runs or overly aggressive model selection. The main contributions are:

1) A dual-axis reliability-decoupling framework that conceptually and quantitatively disentangles confidence from uncertainty, clarifying and mitigating overconfidence and confirmation bias inherent in pseudo-labeling.

2) A tri-space calibration method that unifies feature alignment, predictive consensus, and structural perturbation via a unified reliability engine, ensuring pseudo-labels are both semantically consistent and structurally robust.

3) We advocate a more rigorous evaluation protocol using multi-run results for both best and final checkpoints, challenging the prevalence of optimistic, single-run reporting.

\begin{figure*}[t]
\includegraphics[width=0.98\linewidth]{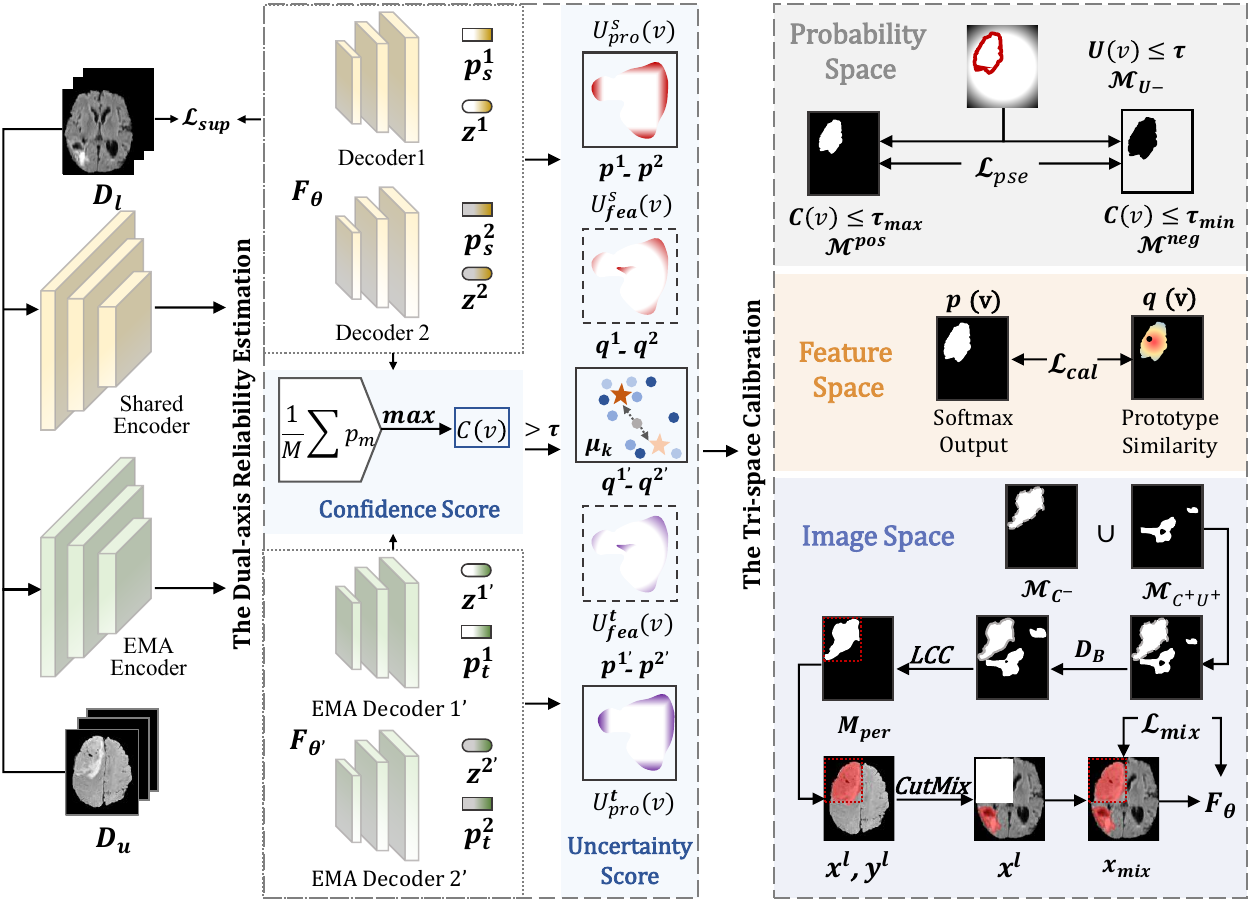}
\centering
\caption{Overview of our proposed TCSeg framework.}
\label{fig:fig2}
\end{figure*}

\section{Related Work}

\subsection{Semi-Supervised Medical Image Segmentation}

The central challenge in SSL is the extraction of high-fidelity supervisory signals from unlabeled data \cite{jing2020self}. Prior approaches diverge in deriving these signals, broadly bifurcating into consistency regularization \cite{luo2021semi,yu2019uncertainty,qiao2022semi} and pseudo-labeling \cite{9761681,9757875,qiu2023federated} paradigms. Consistency regularization posits that the model should yield invariant predictions under perturbations, pseudo-labeling operates on a self-training basis, where the model iteratively generates explicit labels for unlabeled data. These ideas are often combined in teacher-student pipelines, which introduces a well-known vulnerability: when the teacher is essentially a smoothed student, their predictions can become highly correlated, reducing the diversity of supervision and making the framework susceptible to self-reinforcing errors.

\subsection{Reliability Estimation in SSL}

Two main strategies dominate reliability estimation in SSL. Monte Carlo dropout approximates Bayesian uncertainty through repeated stochastic passes \cite{gal2016dropout,yu2019uncertainty,sedai2019uncertainty}, but this becomes prohibitively expensive for 3D volumes. Ensemble methods, using multi-head architectures or co-training, measure predictive disagreement more efficiently \cite{guo2017calibration,li2020transformation,luo2021efficient}, yet still collapse reliability into a single scalar from probability distributions.
In a word, most methods threshold consistency, entropy, variance, or evidential scores, treating reliability as one-dimensional. For instance, DCNet \cite{chen2023decoupled} attempts to decouple unlabeled regions through dynamic thresholding and directional consistency optimization. However, their reliability modeling still fundamentally relies on a single confidence-consistency abstraction derived from output distributions. This may assign high confidence to incorrect predictions \cite{guo2017calibration}—a miscalibration problem where confident does not mean correct. 
Thus, this work aims to address these limitations by explicitly treating confidence and uncertainty as separate reliability axes rather than conflated scalars, and further modeling reliability jointly across probability, feature, and image spaces.

\section{Method}

\subsection{The Trap of Overconfidence}

Let $\mathcal{D}_l = \{(x_i^l, y_i^l)\}_{i=1}^{N_l}$ and $\mathcal{D}_u = \{x_j^u\}_{j=1}^{N_u}$ denote the labeled and unlabeled sets in the training set. The general segmentation loss can be denoted as:
\begin{equation}
\mathcal{L}_{total} = \mathcal{L}_{sup}(\mathcal{D}_l; x_i^l, y_i^l) + \mathcal{L}_{unsup}(\mathcal{D}_{l+u}; x_i^l, y_i^l, x_i^u).
\end{equation}
Because the loss imposed on unlabeled data is not fully reliable and requires calibration, SSL methods often employ a confidence threshold $\tau$ on the prediction $\tilde{y}=argmax(p(\tilde{v}))$ to gate $\mathcal{L}_{unsup}$:
\begin{equation}
\mathcal{L}_{unsup} = \mathds{1}(R(\tilde{v}) > \tau) \cdot L_{ce}(p(v),\tilde{y}),
\end{equation}
\begin{equation}
    R(v) \approx C(v) := \max_{k} p(y=k|v),
\end{equation}
where $p(v)$ is the predictions, $\mathds{1}$ denotes the indicator, which takes the value of 1 when the condition is satisfied and 0 otherwise. A core premise of this paradigm is that confidence mask serves as a faithful proxy for correctness, which equates to assuming that the posterior probability is perfectly calibrated: 
\begin{equation}
\mathbb{P}(\tilde{y} = y^* \mid \max(p(\tilde{v})) = c) = c, \quad \forall c \in [0, 1],
\end{equation}
where $y^*$ is the unseen label. Since $L_{ce}$ minimizes the KL divergence between predictions and sharp pseudo-labels, it pushes the predictive distribution $p(y|v)$ toward a simplex vertex, collapsing into a Dirac delta distribution:
\begin{equation}
\lim_{t \to \infty} p(y|v; \theta_t) \rightarrow \delta(y - \tilde{y}),
\end{equation}
where $\delta$ is the Kronecker delta. This leads to calibration collapse: the model becomes confidently wrong ($\max(p(v)) \to 1$ even $\tilde{y} \neq y^*$), creating a confirmation bias loop that cannot be detected by confidence alone.

\subsection{The Dual-axis Reliability Estimation}
We assign each voxel a reliability vector $R(v)=\langle C(v),U(v)\rangle$ to distinguish confidence from uncertainty, $U(v)$ is a tuple comprising probability-space and feature-space disagreement $U(v)=\langle U_{pro}(v),U_{fea}(v)\rangle$.

\subsubsection{Confidence Score}
The confidence score is defined as the consensus preference across multiple prediction views, two student decoders and their teacher counterparts. Let $\mathbf{P}_s = \{p_s^1, p_s^2\}$ and $\mathbf{P}_t = \{p_t^1, p_t^2\}$ denote the probabilities from the two student decoders and teacher decoders, respectively. The confidence is denoted as the maximum of the ensemble average:
\begin{equation}
    C(v) = \max_{k} \left( \frac{1}{M} \sum_{m=1}^{M} p_m(k|v) \right),\quad p_m \in \mathbf{P}_s \cup \mathbf{P}_t,
\end{equation}
where $m$ represents the $m$-th prediction view.

\subsubsection{Uncertainty Score}
The uncertainty score captures evidence disagreement from two complementary sources, probability-space divergence between the heterogeneous decoders ($U_{pro}$), and feature-space structural deviation ($U_{fea}$):
\begin{equation}
    U_{pro}^r(v) = \| p_r^1(v) - p_r^2(v) \|_1, \quad r \in {\{s,t\}}.
\end{equation}
\begin{equation}
U_{fea}^r(v) = \| {q}_r^1(v) - {q}_r^2(v) \|_1, \quad r \in {\{s,t\}},
\end{equation}
where $q(v)$ is a prototype-aligned similarity prediction: 
\begin{equation}
\label{eqcos}
{q}(v)
=
\cos\!\left( z_v,\, \mu \right)
= 
\frac{
z_v^{\top}\,\mu
}{
\left\|z_v\right\|_2 \,
\left\|\mu\right\|_2
}.
\end{equation}
where $z_v$ is the feature embedding of voxel $v$, and $\mu$ denotes the class prototype computed from high-confidence voxels. For class $k$, we form a prototype by averaging embeddings within the predicted class and passing the confidence filter:
\begin{equation}
\mu_k = \frac{\sum_{v \in V_k} \mathds{1}(C(v) \geq \tau) \cdot z_v}{\sum_{v \in V_k} \mathds{1}(C(v) \geq \tau) + \epsilon},
\end{equation}
where $V_k = \{v : \arg\max_k p(v) = k\}$, $\epsilon$ is a small constant.

\subsection{The Tri-space Calibration}

Our tri-space calibration uses the decoupled scores $C(v)$ and $U(v)$ to define voxel-wise reliability. Thresholding the two axes partitions unlabeled voxels into four groups: only the $C^+U^-$ set (high-confidence and low-uncertainty) is used for pseudo supervision, while the other regimes should be handled explicitly to prevent confirmation bias. Thus, we enforce calibration across three complementary spaces.

\subsubsection{Probability Space}

We employ a collaborative reliability consensus strategy to delineate reliable pseudo-labels. We first enforce cross-view consistency to get low-uncertainty predictions:
\begin{equation}
\mathcal{M}_{U^-} = \mathds{1}\bigg[\left(
\bigwedge_{r \in \{s,t\}} U_{pro}^r \leq \tau\right)
\ \wedge \
\left(\bigwedge_{r \in \{s,t\}} U_{fea}^r \leq \tau\right)
\bigg],
\end{equation}
where $\tau$ is a consistency tolerance. The upper and lower confidence bounds  $\tau_{\text{max}}$ and $\tau_{\text{min}}$ are then used to identify high-confidence positive and negative supervision signals:
\begin{equation}
\mathcal{M}_{C^+U^-}^{pos}(v) = \mathds{1}(C(v) \geq \tau_{\text{max}}) \cdot \mathcal{M}_{U^-}(v),
\end{equation}
\begin{equation}
\mathcal{M}_{C^+U^-}^{neg}(v) = \mathds{1}(C(v) \leq \tau_{\text{min}}) \cdot \mathcal{M}_{U^-}(v).
\end{equation}

The probability-space calibration loss is formulated as:
\begin{equation}
\begin{aligned}
&\mathcal{L}_{pse} = \frac{1}{|\mathcal{M}_{C^+U^-}^{pos}|} \sum_{v \in \mathcal{D}_{u}} \mathcal{M}_{C^+U^-}^{pos}(v) \cdot \mathcal{L}_{ce}(p(v), \tilde{y}(v)) \\
&+ \frac{1}{|\mathcal{M}_{C^+U^-}^{neg}|} \sum_{v \in \mathcal{D}_{u}} \mathcal{M}_{C^+U^-}^{neg}(v) \cdot \mathcal{L}_{ce}(1 - p(v), 1 - \tilde{y}(v)).
\end{aligned}
\end{equation}

\subsubsection{Feature Space}

To rectify structural outliers, we enforce agreement between probability outputs $p(v)$ and prototype-based semantic affinity $q(v)$. This pulls the latent representations of same-class voxels toward their centroid, ensuring that high confidence is supported by semantic density, filtering predictions that are numerically confident but representationally atypical. The feature calibration loss $\mathcal{L}_{cal}$ thus can be denoted as:

\begin{equation}
\begin{aligned}
\mathcal{L}_{cal} = \frac{1}{|\mathcal{D}_{l+u}|} \sum_{v \in \mathcal{D}_{l+u}}
&\bigg[ \sum_{g\in\{p,q\}} \!\bigl \|g_s^{1}(v)-g_s^{2}(v)\bigr\|^2 \\
& + \sum_{i=1}^{2} \ \bigl \|p_s^{i}(v)-q_s^{i}(v)\bigr\|^2 \bigg].
\end{aligned}
\end{equation}

\subsubsection{Image Space}

Static consistency fails to address cognitive blind spots, particularly in regions characterized by low confidence yet low uncertainty, as well as in over-confident regions. Thus, we propose a dynamic structural perturbation module to focus augmentation on cognitive blind spots. A perturbation target mask $\mathcal{M}_{per}$ is first constructed:
\begin{equation}
\mathcal{M}_{per}
= \mathcal{P}_{LCC}
\big(
\mathcal{D}_{B}(
\mathcal{M}_{C^-} \cup \mathcal{M}_{C^+U^+}
)
\big).
\end{equation}
where $\mathcal{D}_B$ denotes binary dilation and $\mathcal{P}_{LCC}$ projects onto the largest connected component. We then apply a targeted CutMix by randomly replacing unreliable regions, encouraging the network to relearn robust boundary features under strong perturbations:

\begin{equation}
x_{mix} = (1 - \mathcal{M}_{\text{per}}) \odot x^l + \mathcal{M}_{\text{per}} \odot x^u,
\end{equation}

\begin{equation}
\tilde{y}_{mix} = (1 - \mathcal{M}_{\text{per}}) \odot y^l + \mathcal{M}_{\text{per}} \odot \tilde{y}^u_t,
\end{equation}

where $\odot$ denotes element-wise multiplication. The mixed sample is then fed into the model, and the prediction is supervised by the mixed pseudo-label:

\begin{equation}
\mathcal{L}_{mix} = \mathcal{L}_{ce}(p_s(v_{mix}), \tilde{y}_{mix}).
\end{equation}

\subsection{Reliability-driven Overall Optimization}

To disrupt the feedback loop where a model reinforces its own erroneous predictions, we integrate the proposed reliability estimation and tri-space calibration into a reciprocal two-branch learning scheme, where the two decoders alternately serve as supervisors and learners. For labeled sets, both branches are supervised independently:
\begin{equation}
\mathcal{L}_{sup} = \sum_{b \in \{1,2\}} 
\Big(\mathcal{L}_{dice}(\tilde{y}^b, y^l) + \mathcal{L}_{ce}(p^b(v), y^l)\Big).
\end{equation}
For unlabeled sets, branch $b$ is guided by its peer branch $\bar{b}$, and $\mathcal{L}_{pse}$ is redefined as:
\begin{equation}
\begin{aligned}
&\mathcal{L}_{pse} = \frac{1}{|\mathcal{M}_{C^+U^-}^{pos}|} \sum_{v \in \mathcal{D}_{u}} \mathcal{M}_{C^+U^-}^{pos}(v) \cdot \mathcal{L}_{ce}(p^b(v), \tilde{y}^{\bar{b}}(v)) \\
&+ \frac{1}{|\mathcal{M}_{C^+U^-}^{neg}|} \sum_{v \in \mathcal{D}_{u}} \mathcal{M}_{C^+U^-}^{neg}(v) \cdot \mathcal{L}_{ce}(1 - p^b(v), 1 - \tilde{y}^{\bar{b}}(v)).
\end{aligned}
\end{equation}

Similarly, $\mathcal{L}_{mix}$ can be reformulated as:
\begin{equation}
\mathcal{L}_{mix} = \mathcal{L}_{ce}(p_s^b(v_{mix}), \tilde{y}^{\bar{b}}_{mix}).
\end{equation}

Thus, the overall loss can be defined as:
\begin{equation}
\mathcal{L}_{total} = \mathcal{L}_{sup} + \mathcal{L}_{pse} + \mathcal{L}_{cal} +  \mathcal{L}_{mix}
\end{equation}

\begin{table*}[ht]
\caption{Reported results of different methods (DSC \%). Please refer to \textbf{Appendix~\ref{appD}} for more results.}
\centering
\small
\begin{threeparttable}
\begin{tabular}{ccccccccc}
\toprule
\multirow{2.5}{*}{Method} &  \multirow{2.5}{*}{Protocol} & \multicolumn{2}{c}{LA} & \multicolumn{2}{c}{Pancreas-CT} & \multicolumn{2}{c}{BraTS2019\tnote{†}} \\
\cmidrule{3-4} \cmidrule{5-6} \cmidrule{7-8}
&  & 10\% / 90\% & 20\% / 80\% &  10\% / 90\% & 20\% / 80\% &  10\% / 90\% & 20\% / 80\%  \\
\midrule

CauSSL \cite{miao2023caussl}               & last  & /     & /     & 72.89 & 80.92 & 83.54 & /     \\
TAC \cite{chen2022semi}                   & last  & 84.73 & 87.75 & /     & /     & /     & /     \\
DTC \cite{luo2021semi}                    & last  & 87.51 & 89.42 & /     & 78.27 & /     & /     \\
CDS \cite{hu2025competitive}              & last  & 87.21 & 90.33 & /     & / & /     & 86.12     \\
ARCO-SG \cite{you2023rethinking}          & last  & 89.90 & /     & /     & /     & /     & /     \\
MCF \cite{wang2023mcf}                    & last  & /     & 88.71 & /     & 75.00 & /     & /     \\
BSNet \cite{10375506}                     & last  & /     & 90.43 & /     & /     & /     & /     \\
DBiSL \cite{LI2026131439}                & last  & 88.78     & 90.54 & /     & 81.09 & 85.09     & /   \\
SFR \cite{10847777}                       & last  & /     & 91.00 & /     & /     & /     & 86.09 \\
AUA \cite{wu2023exploring}                & last  & /     & \second{91.08} & /     & 79.81 & /     & /     \\
TraCoCo \cite{liu2024translation}         & last  & 89.29 & 90.94 & 79.22 & 81.80 & \second{85.71} & \best{86.69} \\
Our median                                & last  & \second{90.28} & 90.83 & \second{81.08} & \second{83.44} & 85.27 & 86.47 \\
Our maximum                               & last  & \best{90.39} & \best{91.36} & \best{81.20} & \best{84.14} & \best{86.52} & \second{86.68} \\

\midrule
UG-MCL \cite{zhang2023uncertainty}         & best  & /     & 90.16 & /     & /     & 82.82 & 83.61 \\
CPCL \cite{9741294}                       & /     & /     & /     & /     & /     & 83.36 & /     \\
SC-SSL \cite{10328616}                    & best  & /     & /     & 70.09 & 80.76 & /     & /     \\
SGTC \cite{yan2025sgtc}                   & best  & 84.70 & /     & /     & /     & /     & /     \\
AUSS \cite{adiga2024anatomically}         & best  & 86.58 & 88.60 & /     & /     & /     & /     \\
CPL \cite{LV2026112321}                   & best  & 87.01 & /     & /     & /     & /     & /     \\
CRDS \cite{aralikatti2023dual}            & /     & 88.51 & /     & /     & /     & /     & /     \\
AC-MT \cite{xu2023ambiguity}              & best  & 89.12 & 90.31 & /     & /     & 83.77 & 84.63 \\
URPC \cite{luo2022semi}                   & best  & /     & /     & 74.89 & 80.31 & 84.16 & 85.49 \\
BaPC \cite{wang2024boundary}              & best  & 88.55 & 89.71 & 68.64 & 76.36 & /     & /     \\
MRPL \cite{su2024mutual}                  & best  & 89.86 & 91.02 & 75.93 & 81.53 & 84.29 & 85.47 \\
BOA-Net \cite{10422862}                   & /     & 88.91 & 91.38 & 76.06 & 80.80 & 85.43 & 85.88 \\
MC-Net+ \cite{wu2022mutual}               & best  & 88.96 & 91.07 & 74.01 & 80.59 & /     & /     \\
VerSemi \cite{10945994}                   & best  & 89.01 & 90.89 & 78.62 & 83.55 & /     & /     \\
Co-BioNet \cite{peiris2023uncertainty}    & best  & 89.20 & 91.26 & 77.89 & 82.22 & /     & /     \\
ICSSL \cite{10530285}                     & best  & 89.25 & /     & /     & 81.17 & /     & /     \\
BCP \cite{bai2023bidirectional}           & best  & 89.62 & /     & /     & 82.91 & /     & /     \\
TraCoCo \cite{liu2024translation}         & best  & 89.86 & 91.51 & /     & 83.36 & \second{85.71} & \best{86.69} \\
W2SPC \cite{yang2025semi}                 & /     & 90.23 & 91.05 & /     & /     & /     & /     \\
OTCMC \cite{guo2025optimal}               & /     & 90.26 & 90.86 & /     & /     & /     & /     \\
A\&D \cite{wang2023towards}               & best  & 90.31 & /     & /     & /     & /     & /     \\
CC-Net \cite{wang2025data}                & best  & \second{90.82} & \second{91.60} & 81.93 & 83.89 & /     & /     \\
Our median                                & best  & 90.73 & 91.44 & \second{82.22} & \second{83.94} & 85.27 & 86.47 \\
Our maximum                               & best  & \best{90.85} & \best{91.62} & \best{82.52} & \best{84.14} & \best{86.52} & \second{86.68} \\

\bottomrule
\end{tabular}
\begin{tablenotes}[flushleft]
\footnotesize
\item[†] BraTS2019 provides a validation set, thus the results under the two protocols coincide.
\end{tablenotes}
\end{threeparttable}
\label{tab:tabcompare}
\end{table*}

\section{Experiments}
\subsection{Datasets and Backbone Network}
We evaluate on three public datasets: Left Atrium (LA) \cite{XIONG2021101832}, Pancreas-CT (NIH) \cite{roth2015deeporgan}, and BraTS2019 \cite{menze2014multimodal}. All models are implemented using SGD for 20k iterations (lr $=0.01$) with batch size 4 (2 labeled + 2 unlabeled). Further specifics and implementation details are provided in the \textbf{Appendix~\ref{appA}} and \textbf{Appendix~\ref{appB}}. Our model is built on a standard VNet-style backbone, which is consistent with the backbone family used by most prior semi-supervised 3D medical image segmentation methods. It adopts a shared five-stage encoder with hierarchical downsampling and multi-scale skip connections, followed by two parallel decoders. Both decoders share the same encoder and standard decoding structure, while using different upsampling operators to produce complementary outputs and decoder features.

\subsection{Evaluation Protocol}

The semi-supervised medical image segmentation literature exhibits considerable inconsistency in checkpoint selection strategies. Reported performance can vary substantially with the checkpointing rule: some works report best (the peak score observed during training), while others report last (the final checkpoint). We argue that while the best checkpoint represents a model's theoretical peak, the last checkpoint better reflects its convergence stability and deployment viability in real-world scenarios where test labels are unavailable. To address the concerns regarding result overconfidence, we establish a rigorous benchmarking protocol designed to disentangle genuine performance gains from stochastic luck. Specifically, each setting is run five times with different random seeds, and we report both best and last. We summarize the distribution across runs using the median to reflect typical behavior, and also report the maximum to facilitate comparison with prior single-run, best-checkpoint style reporting. This combination exposes instability, reduces sensitivity to randomness, and discourages conclusions driven by a fortunate checkpoint or seed. We evaluate using standard segmentation metrics, DSC, ASD, and 95HD, specifics are provided in the \textbf{Appendix~\ref{appC}}.

\begin{figure*}[t]
\centering
\includegraphics[width=1.0\linewidth]{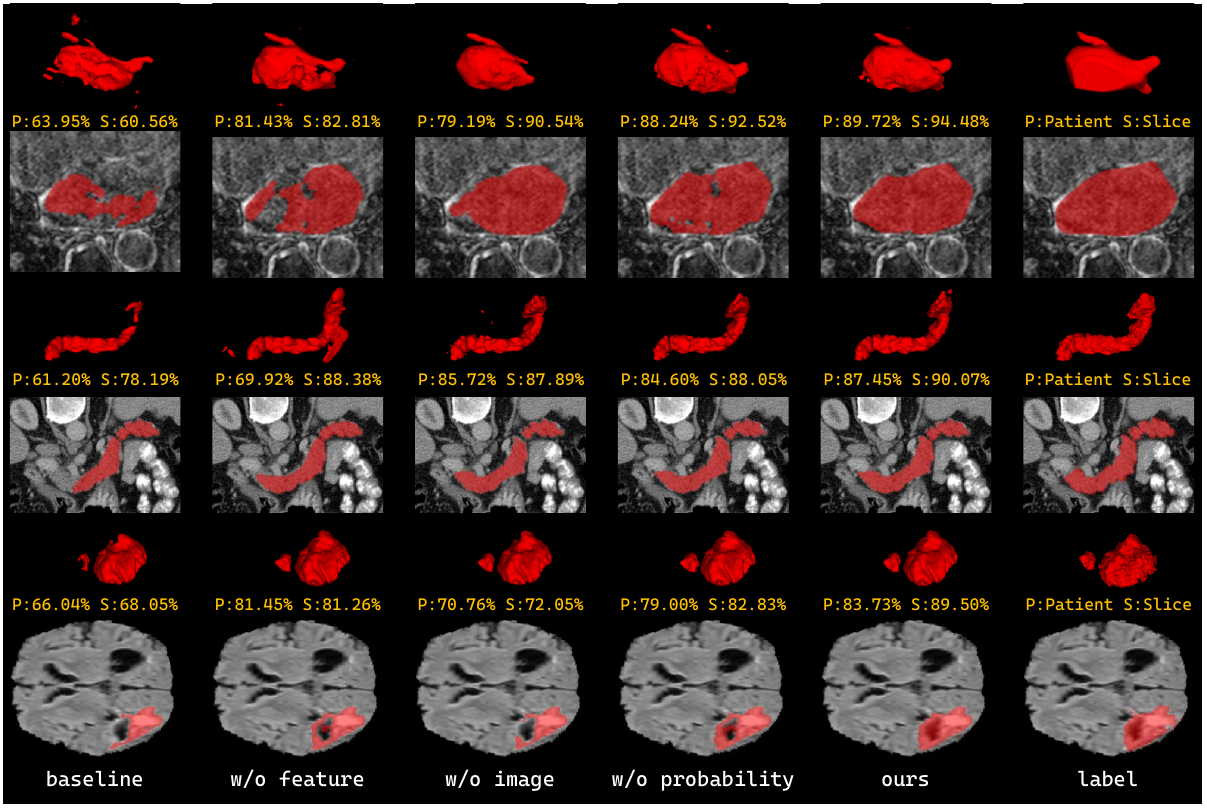}
\caption{Visualization of the segmentation results of different methods. None of the results were post-processed.}
\label{fig:fig5}
\end{figure*}

\subsection{Comparison with SOTA Methods}

Table~\ref{tab:tabcompare} benchmarks the proposed TCSeg against recent state-of-the-art methods on the LA, Pancreas-CT, and BraTS2019 datasets. To ensure a transparent comparison, we explicitly categorize results based on the evaluation protocol (last vs. best). Under the rigorous last protocol—which reflects genuine convergence stability without oracle checkpoint selection—our framework demonstrates a substantial performance margin. For instance, on the challenging 10\% labeled Pancreas-CT task, our median DSC of 81.08\% surpasses the nearest competitor (TraCoCo) by approximately 1.8\%. It shows that TCSeg consistently maintains stable convergence across repeated runs, indicating the proposed tri-space reliability calibration effectively mitigates confirmation bias accumulation during training. 
The performance gaps under the optimistic best-checkpoint protocol appear numerically compressed because many recent methods already rely on best-checkpoint selection, which elevates the reported peak performance. As a result, the remaining improvement margin becomes relatively limited on mature public benchmarks. In this setting, even modest absolute gains are meaningful.
Crucially, our median performance consistently rivals or exceeds the maximum scores reported by competitive methods like CC-Net and VerSemi. This highlights the observed gains are not due to a single favorable initialization but reflect a stable and reproducible improvement, thus indicating that the intrinsic robustness of our approach against initialization noise.

\subsection{Ablation Studies on Dual-axis Assessment}

To validate the necessity of disentangling confidence and uncertainty, we conduct an ablation study on the proposed dual-axis reliability assessment in Table~\ref{tab:tabdual}. Removing either the uncertainty branch (w/o $U(v)$) or the confidence branch (w/o $C(v)$) consistently degrades performance across all datasets. Although the numerical gaps appear moderate on mature public benchmarks, both single-axis variants still retain partial reliability cues and therefore remain stronger than other naive pseudo-label selection methods (Table~\ref{tab:tabcompare}). In contrast, the proposed dual-axis strategy consistently achieves the best performance, improving the mean DSC to 86.23\%. These results suggest that confidence and uncertainty provide complementary reliability signals, and jointly modeling both leads to more robust pseudo-label selection and optimization stability.

\subsection{Ablation Studies on Three Spaces}

Another ablation study is reported in Table~\ref{tab:tab3} to assess the contribution of each calibration space. When all three components are enabled, the model attains the highest mean DSC of 86.23\%. Removing the probability-space calibration (w/o prob.) leads to a moderate drop to 85.13\%, indicating that confidence-band filtering and probability-space consensus help stabilize pseudo-label selection. Excluding the image-space structural perturbation (w/o img.) further reduces the mean DSC to 84.00\% and consistently affects all datasets, suggesting that reliability-guided perturbations improve structural robustness across varying anatomical distributions and decision boundaries.
Removing feature-space prototypes (w/o feat.) produces the most severe degradation on Pancreas-CT, where the performance drops from 81.08\%/83.44\% to 59.28\%/69.23\%. This suggests that feature-space prototype alignment is particularly important for low-contrast and boundary-ambiguous anatomical structures such as the pancreas. This observation should not be interpreted as implying a universally fixed importance ranking among the three spaces. In contrast, image-space perturbation exhibits more consistent benefits across all datasets, while the relative contribution of each component appears to depend on dataset characteristics such as local contrast, boundary ambiguity, and anatomical variability. Overall, the probability-, feature-, and image-space calibrations act as complementary mechanisms that jointly contribute to robust semi-supervised segmentation performance.

\begin{figure}[t]
\centering
\includegraphics[width=1.0\columnwidth]{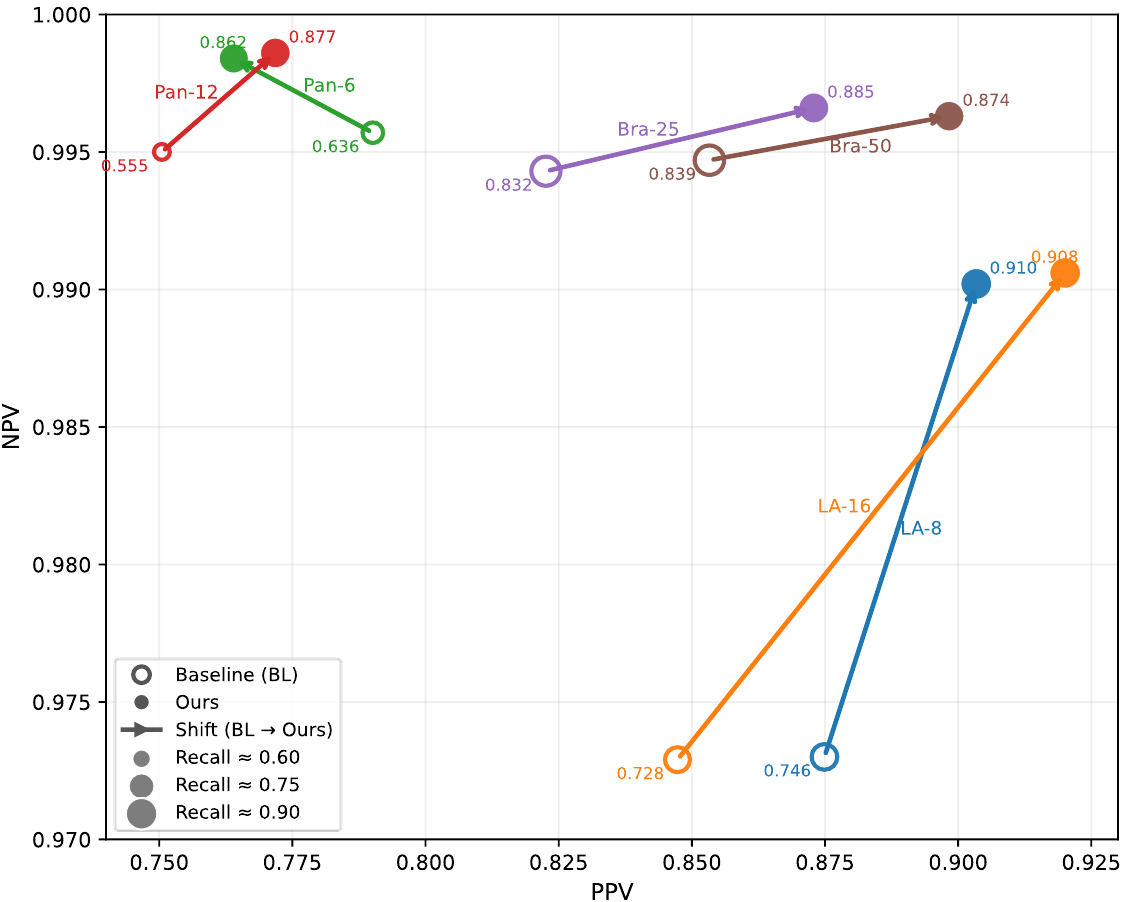}
\caption{
PPV--NPV analysis of pseudo labels on the unlabeled training set. Each pair of hollow (baseline) and filled (ours) circles corresponds to one dataset and labeled ratio. The horizontal and vertical axes report the positive predictive value and negative predictive value, respectively, while the marker size encodes recall. Arrows indicate the shift from the baseline model to the proposed method under the same setting. PPV and NPV are computed as $\mathrm{PPV} = \frac{\mathrm{TP}}{\mathrm{TP} + \mathrm{FP}}$ and $\mathrm{NPV} = \frac{\mathrm{TN}}{\mathrm{TN} + \mathrm{FN}}$, respectively.
}
\label{fig:figshift}
\end{figure}

\subsection{Confidence Analysis on Unlabeled Samples}

Fig.~\ref{fig:figshift} summarizes how the proposed tri-space calibration affects the reliability of pseudo labels on unlabeled training volumes. 
We extract the segmentation masks predicted for all unlabeled training data (i.e., the pseudo-labels used during training), and compute the Positive Predictive Value (PPV), Negative Predictive Value (NPV), and Recall based on the resulting 3D segmentation volumes. 
For all six dataset and labeling configurations, the arrows consistently move towards the upper-right region and are accompanied by larger markers, indicating that our method achieves higher or comparable PPV together with uniformly improved NPV and substantially increased recall. The gains are particularly pronounced for the abdominal tasks (Pan-6 and Pan-12), where recall rises from roughly 0.56–0.64 to above 0.86 while NPV remains extremely high, suggesting that many previously missed foreground voxels—often associated with over-confident background predictions near ambiguous boundaries—are now correctly recovered without sacrificing the reliability of negative predictions. For the LA and BraTS settings, PPV, NPV, and recall all improve simultaneously. Taken together, these shifts show that the proposed reliability-driven calibration does not simply inflate confidence, but instead yields pseudo labels that are both more complete and better aligned with the true error statistics on unlabeled data, thereby alleviating the over-confidence behavior observed in standard semi-supervised baselines.

\begin{figure}[t]
\centering
\includegraphics[width=1.0\columnwidth]{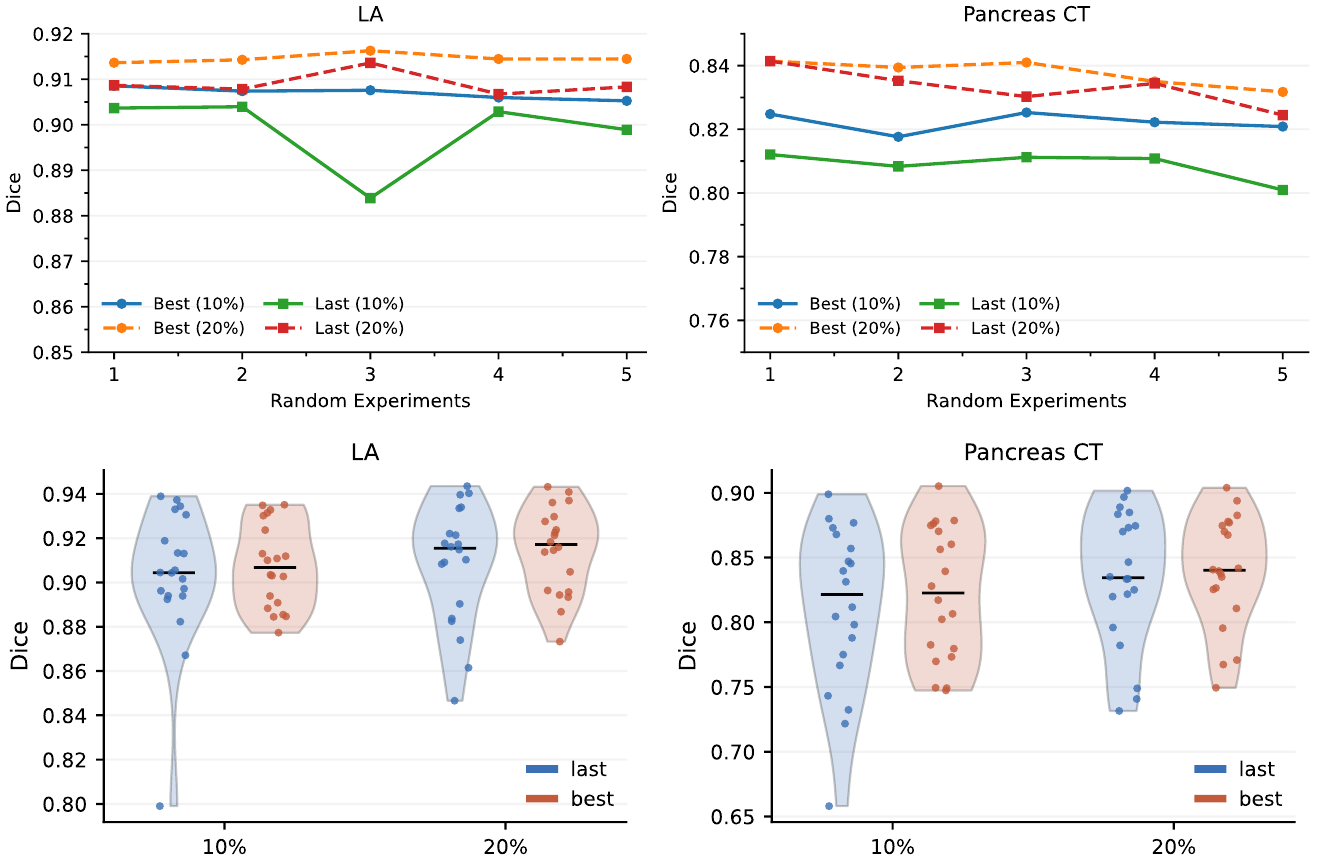}
\caption{Run-to-run performance under different checkpointing strategies on LA and Pancreas-CT (BraTS2019 provides a validation split, thus the best and last results are treated as equivalent). The first row shows the performance of the two protocols across five repeated experiments, while the second row displays the distribution of segmentation results obtained by both protocols. Under the last-checkpoint protocol, the five-run mean$\pm$std results are: LA 89.86$\pm$0.12 (10\%) / 90.90$\pm$0.09 (20\%), and Pancreas-CT 80.86$\pm$0.28 (10\%) / 83.31$\pm$0.37 (20\%).}

\label{fig:figseeds}
\end{figure}

\begin{table}[t]
\caption{Ablation studies on Dual-axis Assessment (DSC:\%).}
\centering
\small
\resizebox{\columnwidth}{!}{%
\begin{tabular}{ccccccccc}
\toprule
\multirow{2.5}{*}{Method} & \multicolumn{2}{c}{LA} & \multicolumn{2}{c}{Pancreas-CT} & \multicolumn{2}{c}{BraTS2019} & \multirow{2.5}{*}{Mean} \\
\cmidrule{2-3} \cmidrule{4-5} \cmidrule{6-7}
& 10\% & 20\% & 10\% & 20\% & 10\% & 20\% & \\
\midrule
w/o $U(v)$  & 90.12 & 90.72 & 80.42 & 82.00 & 84.94 & 85.88 & 85.68 \\
w/o $C(v)$ & 88.88 & 90.76 & 78.82 & 81.93 & 84.59 & 86.22 & 85.20 \\
Dual-axis & 90.28 & 90.83 & 81.08 & 83.44 & 85.27 & 86.47 & 86.23 \\
\bottomrule
\end{tabular}%
}
\label{tab:tabdual}
\end{table}

\begin{table}[t]
\caption{Ablation studies on three spaces (DSC:\%).}
\centering
\small
\resizebox{\columnwidth}{!}{%
\begin{tabular}{ccccccccc}
\toprule
\multirow{2.5}{*}{Method} & \multicolumn{2}{c}{LA} & \multicolumn{2}{c}{Pancreas-CT} & \multicolumn{2}{c}{BraTS2019} & \multirow{2.5}{*}{Mean} \\
\cmidrule{2-3} \cmidrule{4-5} \cmidrule{6-7}
& 10\% & 20\% & 10\% & 20\% & 10\% & 20\% & \\
\midrule
Only sup & 76.18 & 82.36 & 55.72 & 65.60 & 76.29 & 80.01 & 72.69 \\
w/o prob. & 89.84 & 90.27 & 78.65 & 80.59 & 85.63 & 85.80 & 85.13 \\
w/o img.  & 87.68 & 90.62 & 75.48 & 81.35 & 84.07 & 84.81 & 84.00 \\
w/o feat. & 89.61 & 90.80 & 59.28 & 69.23 & 85.57 & 86.06 & 80.09 \\
Ours & 90.28 & 90.83 & 81.08 & 83.44 & 85.27 & 86.47 & 86.23 \\
\bottomrule
\end{tabular}%
}
\label{tab:tab3}
\end{table}

\subsection{Why Need Multi-run and Different Protocols}

SSL conclusions can hinge on two largely orthogonal choices: the random seed and the checkpointing rule. Even within the same dataset and label budget, the best checkpoint can look deceptively stable while the last checkpoint---which matches a straightforward train-and-deploy workflow---exhibits a noticeably larger spread. 
Such gaps indicate that a method can appear strong after favorable checkpoint selection yet deliver a noticeably weaker final model, overstating the reliability one would expect in practice.
To reflect this reality, we adopt a multi-run protocol and report both best and last with the median and maximum across runs: the median captures the typical behavior one should expect, while the maximum retains an upper-bound reference, together reducing the chance that comparisons are driven by outliers or checkpoint cherry-picking.
Fig.~\ref{fig:figseeds} indicated that our method keeps best and last closely aligned across seeds, with a pronounced discrepancy appearing only once, suggesting a more stable optimization.

\subsection{Parameter Sensitivity Analysis}
\begin{figure}[t]
\centering
\includegraphics[width=1.0\columnwidth]{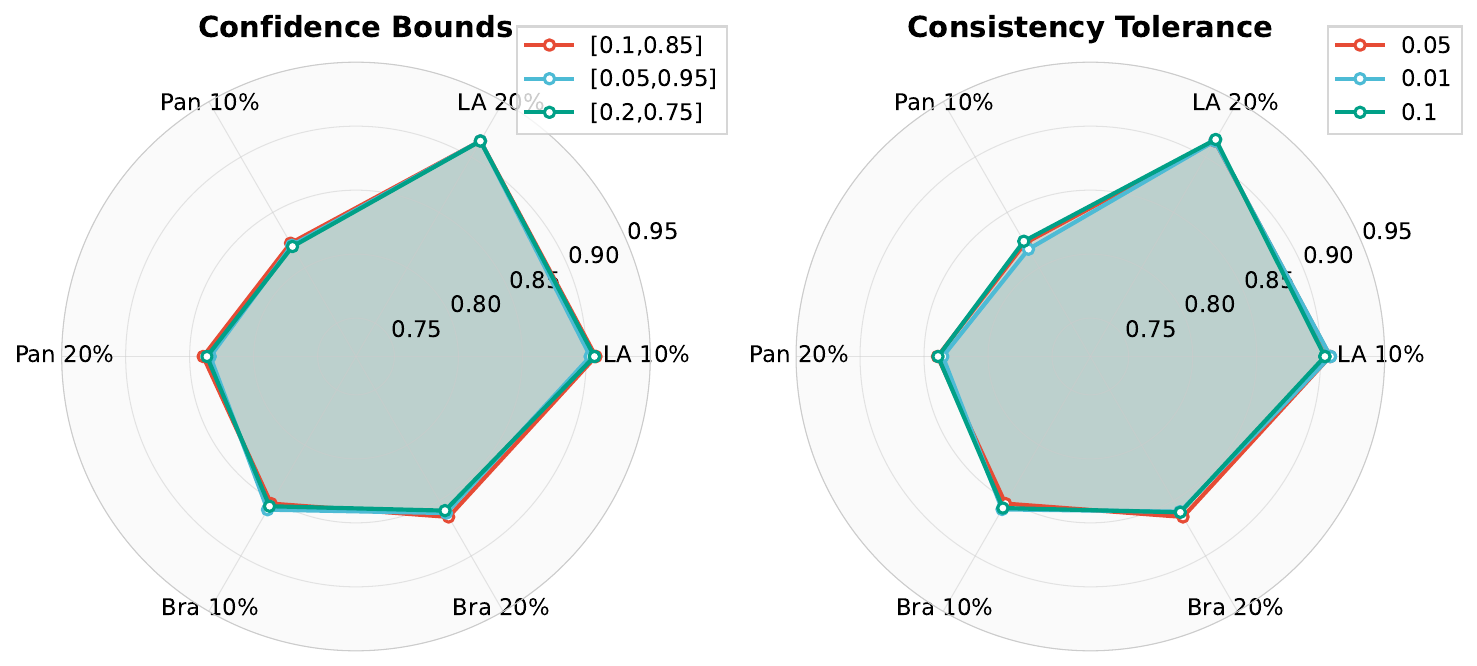}
\caption{Sensitivity analysis with respect to confidence bounds and consistency tolerance. The radius encodes the DSC, and each ray corresponds to one dataset–labeling pair.}
\label{fig:figradar}
\end{figure}

We assess sensitivity to two key hyperparameters by varying the confidence band ($\tau_{min}$,$\tau_{max}$) for pseudo-label selection and the tolerance ($\tau$) used in consistency regularization (Fig.~\ref{fig:figradar}). Across all settings, the radar curves are tightly clustered and largely overlap. Adjusting the tolerance from $0.01$ to $0.10$ and shifting the confidence bounds from $[0.1,0.85]$ to either $[0.05,0.95]$ or $[0.2,0.75]$ changes DSC only marginally, without a consistent trend. These results suggest that the tri-space calibration is stable over a reasonable parameter range, and the gains are not driven by a carefully tuned setting.

\subsection{Computational Efficiency Analysis}

Table~\ref{tab:tabcom} summarizes the computational costs of representative SSL methods on BraTS2019. Our method remains lightweight, with 12.34M parameters, which is comparable to MRPL (12.35M) and notably fewer than CauSSL (23.60M). The training cost is also moderate (0.421~s/iter), well within the range of common SSL pipelines and far below heavier designs such as CC-Net (2.934~s/iter). For completeness, the test-time latency is measured with the full implementation. In fact, all auxiliary modules can be discarded and only the backbone path is retained in deployment, so the actual test-time efficiency can be made fully identical to the backbone. Overall, these results indicate that the performance gains are achieved without introducing excessive computational overhead, while preserving deployment-friendly inference efficiency.

\begin{table}[t]
\centering
\small
\caption{Computational costs of different methods on BraTS2019 dataset (Mem: GPU memory usage during training). All results were obtained using the NVIDIA RTX 4080 GPU.}
\resizebox{\columnwidth}{!}{%
\begin{tabular}{lcccc}
\toprule
Method & \makecell{Params \\ (M)} & \makecell{Train \\ (s/iter)} & \makecell{Test \\ (s/case)}  & \makecell{Mem \\ (GB)} \\
\midrule
DTC \cite{luo2021semi} & 9.45 & 0.645 & 1.02 & 4.62 \\
CauSSL \cite{miao2023caussl} & 23.60 & 0.442 & 1.75 & 8.43 \\
MRPL \cite{su2024mutual}  & 12.35 & 0.334 & 1.67 & 10.40 \\
CC-Net \cite{wang2025data} & 14.79 & 2.934 & 3.03 & 13.10 \\
Ours & 12.34 & 0.421  & 1.66 & 10.83 \\
\bottomrule
\end{tabular}
}
\label{tab:tabcom}
\end{table}

\subsection{Limitations and Future Works}

our reliability analysis is currently limited to semi-supervised 3D medical image segmentation benchmarks and should not be interpreted as establishing universal uncertainty calibration or out-of-distribution robustness. In fact, benchmark improvements do not guarantee safe performance under distribution shift (e.g., new scanners, protocols, or modalities). Moreover, stability on public datasets should not be conflated with clinical readiness, and premature deployment without prospective validation remains a risk. Future work should prioritize multi-site evaluation and calibration assessment under deployment-like conditions. Another limitation is that our framework currently relies on fixed thresholds, and developing more robust adaptive parameter selection strategies may further improve model robustness and reliability. Finally, many prior SSL methods are reported under inconsistent postprocessing, checkpoint-selection, and evaluation protocols. To address this issue, we are currently conducting a benchmark study aimed at systematically re-implementing and re-evaluating more than 25 SSL methods under a unified multi-run protocol, and we plan to report these rigorous standardized comparisons in future works.

\section{Conclusion}
This paper asks a practical question that is often overlooked in semi-supervised 3D medical image segmentation: are we overconfident not only in our models, but also in the results we report? On the modeling side, many SSL pipelines implicitly equate a high softmax score with pseudo-label reliability, which can turn miscalibration into a confirmation-bias loop---confident mistakes are recycled as supervision and become increasingly difficult to correct. On the reporting side, the community has largely converged to single-run, best-checkpoint reporting, frequently selected by monitoring the test set. While this practice can showcase a method's peak performance, it may also exaggerate real-world utility, where deployment typically relies on the final model obtained from a straightforward training run. To address both, we propose a reliability-driven framework that decouples confidence from uncertainty and applies a tri-space calibration mechanism to suppress confidently wrong supervision. Experiments on LA, Pancreas-CT, and BraTS2019 show consistent gains under both best- and last-checkpoint protocols, and multi-run reporting (median and extrema) provides a more informative view of robustness. We hope this perspective encourages reliability-aware semi-supervision and evaluation practices that better reflect deployment.

\section*{Acknowledgements}
This work was supported by the National Natural Science Foundation of China under Grant 62401481, Natural Science Foundation of Sichuan Province under Grant 2025ZNSFSC1450, Fundamental Research Funds for the Central Universities under Grant 2682026ZTPY027, China Postdoctoral Science Foundation under Grant 2024M752683.

\section*{Impact Statement}

This paper aims to improve the reliability of semi-supervised 3D medical image segmentation. Beyond proposing a training framework that reduces confirmation bias in pseudo labeling, another central contribution is a more rigorous evaluation protocol that better match how segmentation systems are used in practice. In much of the recent literature, performance is often summarized by the single best checkpoint from a single run---frequently selected with direct feedback from the test set. While such numbers can be informative as an upper bound, they may paint an overly optimistic picture of deployment behavior, where clinicians typically rely on a final trained model and expect consistent performance across runs, scanners, and patient cohorts. By reporting both last- and best-checkpoint results over repeated runs, and summarizing variability rather than only peaks, our study encourages a shift toward transparency and robustness as first-class objectives. We believe this change is particularly relevant for clinical practice, where unstable or selectively reported gains can lead to inappropriate expectations and hinder safe application.

\nocite{langley00}

\bibliography{example_paper}

@inproceedings{yu2019uncertainty,
  title={Uncertainty-aware self-ensembling model for semi-supervised 3D left atrium segmentation},
  author={Yu, Lequan and Wang, Shujun and Li, Xiaomeng and Fu, Chi-Wing and Heng, Pheng-Ann},
  booktitle={Medical image computing and computer assisted intervention--MICCAI 2019: 22nd international conference, Shenzhen, China, October 13--17, 2019, proceedings, part II 22},
  pages={605--613},
  year={2019},
  organization={Springer}
}

@article{LI2026131439,
title = {Fully differentiable bidirectional dual-task synergistic learning for semi-supervised 3D medical image segmentation},
journal = {Expert Systems with Applications},
volume = {313},
pages = {131439},
year = {2026},
doi = {https://doi.org/10.1016/j.eswa.2026.131439},
author = {Jun Li},
}

@article{guo2025optimal,
  title={Optimal Transport and Central Moment Consistency Regularization for Semi-supervised Medical Image Segmentation},
  author={Guo, Xiuzhen and Yu, Lianyuan and Shi, Ji and Wang, Hongxiao and Zhao, Jiangyuan and Zhang, Rongguo and Li, Hongwei and Lei, Na},
  journal={IEEE Transactions on Medical Imaging},
  year={2025},
  publisher={IEEE}
}

@article{hu2025competitive,
  title={Competitive dual-students using bi-level contrastive learning for semi-supervised medical image segmentation},
  author={Hu, Gang and Zhao, Feng and Houssein, Essam H},
  journal={Engineering Applications of Artificial Intelligence},
  volume={144},
  pages={110082},
  year={2025},
  publisher={Elsevier}
}

@article{menze2014multimodal,
  title={The multimodal brain tumor image segmentation benchmark (BRATS)},
  author={Menze, Bjoern H and Jakab, Andras and Bauer, Stefan and Kalpathy-Cramer, Jayashree and Farahani, Keyvan and Kirby, Justin and Burren, Yuliya and Porz, Nicole and Slotboom, Johannes and Wiest, Roland and others},
  journal={IEEE transactions on medical imaging},
  volume={34},
  number={10},
  pages={1993--2024},
  year={2014},
  publisher={IEEE}
}

@article{wang2025data,
  title={Data augmentation strategies for semi-supervised medical image segmentation},
  author={Wang, Jiahui and Ruan, Dongsheng and Li, Yang and Wang, Zefeng and Wu, Yongquan and Tan, Tao and Yang, Guang and Jiang, Mingfeng},
  journal={Pattern Recognition},
  volume={159},
  pages={111116},
  year={2025},
  publisher={Elsevier}
}

@article{wang2023towards,
  title={Towards generic semi-supervised framework for volumetric medical image segmentation},
  author={Wang, Haonan and Li, Xiaomeng},
  journal={Advances in Neural Information Processing Systems},
  volume={36},
  pages={1833--1848},
  year={2023}
}

@inproceedings{bai2023bidirectional,
  title={Bidirectional copy-paste for semi-supervised medical image segmentation},
  author={Bai, Yunhao and Chen, Duowen and Li, Qingli and Shen, Wei and Wang, Yan},
  booktitle={Proceedings of the IEEE/CVF conference on computer vision and pattern recognition},
  pages={11514--11524},
  year={2023}
}

@article{yang2025semi,
  title={Semi-supervised medical image segmentation via weak-to-strong perturbation consistency and edge-aware contrastive representation},
  author={Yang, Yang and Sun, Guoying and Zhang, Tong and Wang, Ruixuan and Su, Jingyong},
  journal={Medical Image Analysis},
  volume={101},
  pages={103450},
  year={2025},
  publisher={Elsevier}
}

@ARTICLE{10530285,
  author={Huang, Wei and Zhang, Lei and Wang, Zizhou and Wang, Lituan},
  journal={IEEE Transactions on Medical Imaging}, 
  title={Exploring Inherent Consistency for Semi-Supervised Anatomical Structure Segmentation in Medical Imaging}, 
  year={2024},
  volume={43},
  number={11},
  pages={3731-3741},
  doi={10.1109/TMI.2024.3400840}}

@ARTICLE{10945994,
  author={Zeng, Qingjie and Xie, Yutong and Lu, Zilin and Lu, Mengkang and Wu, Yicheng and Xia, Yong},
  journal={IEEE Transactions on Medical Imaging}, 
  title={Segment Together: A Versatile Paradigm for Semi-Supervised Medical Image Segmentation}, 
  year={2025},
  volume={44},
  number={7},
  pages={2948-2959},
  doi={10.1109/TMI.2025.3556310}}

@article{peiris2023uncertainty,
  title={Uncertainty-guided dual-views for semi-supervised volumetric medical image segmentation},
  author={Peiris, Himashi and Hayat, Munawar and Chen, Zhaolin and Egan, Gary and Harandi, Mehrtash},
  journal={Nature Machine Intelligence},
  volume={5},
  number={7},
  pages={724--738},
  year={2023},
  publisher={Nature Publishing Group UK London}
}

@article{wu2022mutual,
  title={Mutual consistency learning for semi-supervised medical image segmentation},
  author={Wu, Yicheng and Ge, Zongyuan and Zhang, Donghao and Xu, Minfeng and Zhang, Lei and Xia, Yong and Cai, Jianfei},
  journal={Medical Image Analysis},
  volume={81},
  pages={102530},
  year={2022},
  publisher={Elsevier}
}

@ARTICLE{10422862,
  author={Li, Congcong and Zhang, Jinshuo and Niu, Dongmei and Zhao, Xiuyang and Yang, Bo and Zhang, Caiming},
  journal={IEEE Transactions on Artificial Intelligence}, 
  title={Boundary-Aware Uncertainty Suppression for Semi-Supervised Medical Image Segmentation}, 
  year={2024},
  volume={5},
  number={8},
  pages={4074-4086},
  doi={10.1109/TAI.2024.3359576}}

@article{su2024mutual,
  title={Mutual learning with reliable pseudo label for semi-supervised medical image segmentation},
  author={Su, Jiawei and Luo, Zhiming and Lian, Sheng and Lin, Dazhen and Li, Shaozi},
  journal={Medical Image Analysis},
  volume={94},
  pages={103111},
  year={2024},
  publisher={Elsevier}
}

@article{aralikatti2023dual,
  title={A dual-stage semi-supervised pre-training approach for medical image segmentation},
  author={Aralikatti, Rajath C and Pawan, SJ and Rajan, Jeny},
  journal={IEEE Transactions on Artificial Intelligence},
  volume={5},
  number={2},
  pages={556--565},
  year={2023},
  publisher={IEEE}
}

@inproceedings{yan2025sgtc,
  title={SGTC: Semantic-guided triplet co-training for sparsely annotated semi-supervised medical image segmentation},
  author={Yan, Ke and Cai, Qing and Zhang, Fan and Cao, Ziyan and Liu, Zhi},
  booktitle={Proceedings of the AAAI Conference on Artificial Intelligence},
  volume={39},
  number={9},
  pages={9112--9120},
  year={2025}
}

@article{LV2026112321,
title = {Clustering-guided contrastive prototype learning: Towards semi-supervised medical image segmentation},
journal = {Pattern Recognition},
volume = {171},
pages = {112321},
year = {2026},
issn = {0031-3203},
doi = {https://doi.org/10.1016/j.patcog.2025.112321},
author = {Zihe Lv and Zhengda Wu and Jinghua Zhu}
}

@ARTICLE{10328616,
  author={Miao, Juzheng and Zhou, Si-Ping and Zhou, Guang-Quan and Wang, Kai-Ni and Yang, Meng and Zhou, Shoujun and Chen, Yang},
  journal={IEEE Transactions on Medical Imaging}, 
  title={SC-SSL: Self-Correcting Collaborative and Contrastive Co-Training Model for Semi-Supervised Medical Image Segmentation}, 
  year={2024},
  volume={43},
  number={4},
  pages={1347-1364},
  doi={10.1109/TMI.2023.3336534}}

@ARTICLE{9741294,
  author={Xu, Zhe and Wang, Yixin and Lu, Donghuan and Yu, Lequan and Yan, Jiangpeng and Luo, Jie and Ma, Kai and Zheng, Yefeng and Tong, Raymond Kai-yu},
  journal={IEEE Journal of Biomedical and Health Informatics}, 
  title={All-Around Real Label Supervision: Cyclic Prototype Consistency Learning for Semi-Supervised Medical Image Segmentation}, 
  year={2022},
  volume={26},
  number={7},
  pages={3174-3184},
  doi={10.1109/JBHI.2022.3162043}}

@ARTICLE{10847777,
  author={Li, Shumeng and Qi, Lei and Yu, Qian and Huo, Jing and Shi, Yinghuan and Gao, Yang},
  journal={IEEE Transactions on Medical Imaging}, 
  title={Stitching, Fine-Tuning, and Re-Training: A SAM-Enabled Framework for Semi-Supervised 3D Medical Image Segmentation}, 
  year={2025},
  volume={44},
  number={10},
  pages={3909-3923},
  doi={10.1109/TMI.2025.3532084}}

@article{you2023rethinking,
  title={Rethinking semi-supervised medical image segmentation: A variance-reduction perspective},
  author={You, Chenyu and Dai, Weicheng and Min, Yifei and Liu, Fenglin and Clifton, David and Zhou, S Kevin and Staib, Lawrence and Duncan, James},
  journal={Advances in neural information processing systems},
  volume={36},
  pages={9984--10021},
  year={2023}
}

@ARTICLE{10375506,
  author={He, Along and Li, Tao and Yan, Juncheng and Wang, Kai and Fu, Huazhu},
  journal={IEEE Transactions on Medical Imaging}, 
  title={Bilateral Supervision Network for Semi-Supervised Medical Image Segmentation}, 
  year={2024},
  volume={43},
  number={5},
  pages={1715-1726},
  doi={10.1109/TMI.2023.3347689}}

@ARTICLE{liu2024translation,
  author={Liu, Yuyuan and Tian, Yu and Wang, Chong and Chen, Yuanhong and Liu, Fengbei and Belagiannis, Vasileios and Carneiro, Gustavo},
  journal={IEEE Transactions on Medical Imaging}, 
  title={Translation Consistent Semi-Supervised Segmentation for 3D Medical Images}, 
  year={2025},
  volume={44},
  number={2},
  pages={952-968},
  doi={10.1109/TMI.2024.3468896}}

@inproceedings{roth2015deeporgan,
  title={Deeporgan: Multi-level deep convolutional networks for automated pancreas segmentation},
  author={Roth, Holger R and Lu, Le and Farag, Amal and Shin, Hoo-Chang and Liu, Jiamin and Turkbey, Evrim B and Summers, Ronald M},
  booktitle={International conference on medical image computing and computer-assisted intervention},
  pages={556--564},
  year={2015},
  organization={Springer}
}

@inproceedings{sedai2019uncertainty,
  title={Uncertainty guided semi-supervised segmentation of retinal layers in OCT images},
  author={Sedai, Suman and Antony, Bhavna and Rai, Ravneet and Jones, Katie and Ishikawa, Hiroshi and Schuman, Joel and Gadi, Wollstein and Garnavi, Rahil},
  booktitle={International conference on medical image computing and computer-assisted intervention},
  pages={282--290},
  year={2019},
  organization={Springer}
}

@inproceedings{gal2016dropout,
  title={Dropout as a bayesian approximation: Representing model uncertainty in deep learning},
  author={Gal, Yarin and Ghahramani, Zoubin},
  booktitle={international conference on machine learning},
  pages={1050--1059},
  year={2016},
  organization={PMLR}
}

@ARTICLE{10982257,
  author={Li, Jun and Zhang, Yijue and Shi, Haibo and Li, Minhong and Li, Qiwei and Qian, Xiaohua},
  journal={IEEE Transactions on Medical Imaging}, 
  title={A Dual-Task Synergy-Driven Generalization Framework for Pancreatic Cancer Segmentation in CT Scans}, 
  year={2025},
  volume={44},
  number={9},
  pages={3783-3794},
  doi={10.1109/TMI.2025.3566376}}

@inproceedings{guo2017calibration,
  title={On calibration of modern neural networks},
  author={Guo, Chuan and Pleiss, Geoff and Sun, Yu and Weinberger, Kilian Q},
  booktitle={International conference on machine learning},
  pages={1321--1330},
  year={2017},
  organization={PMLR}
}

@inproceedings{luo2021semi,
  title={Semi-supervised medical image segmentation through dual-task consistency},
  author={Luo, Xiangde and Chen, Jieneng and Song, Tao and Wang, Guotai},
  booktitle={Proceedings of the AAAI conference on artificial intelligence},
  volume={35},
  number={10},
  pages={8801--8809},
  year={2021}
}

@inproceedings{luo2021efficient,
  title={Efficient semi-supervised gross target volume of nasopharyngeal carcinoma segmentation via uncertainty rectified pyramid consistency},
  author={Luo, Xiangde and Liao, Wenjun and Chen, Jieneng and Song, Tao and Chen, Yinan and Zhang, Shichuan and Chen, Nianyong and Wang, Guotai and Zhang, Shaoting},
  booktitle={Medical Image Computing and Computer Assisted Intervention--MICCAI 2021: 24th International Conference, Strasbourg, France, September 27--October 1, 2021, Proceedings, Part II 24},
  pages={318--329},
  year={2021},
  organization={Springer}
}

@article{shi2021inconsistency,
  title={Inconsistency-aware uncertainty estimation for semi-supervised medical image segmentation},
  author={Shi, Yinghuan and Zhang, Jian and Ling, Tong and Lu, Jiwen and Zheng, Yefeng and Yu, Qian and Qi, Lei and Gao, Yang},
  journal={IEEE transactions on medical imaging},
  volume={41},
  number={3},
  pages={608--620},
  year={2021},
  publisher={IEEE}
}

@article{adiga2024anatomically,
  title={Anatomically-aware uncertainty for semi-supervised image segmentation},
  author={Adiga, Sukesh and Dolz, Jose and Lombaert, Herve},
  journal={Medical Image Analysis},
  volume={91},
  pages={103011},
  year={2024},
  publisher={Elsevier}
}

@article{zhang2023uncertainty,
  title={Uncertainty-guided mutual consistency learning for semi-supervised medical image segmentation},
  author={Zhang, Yichi and Jiao, Rushi and Liao, Qingcheng and Li, Dongyang and Zhang, Jicong},
  journal={Artificial Intelligence in Medicine},
  volume={138},
  pages={102476},
  year={2023},
  publisher={Elsevier}
}

@article{li2020transformation,
  title={Transformation-consistent self-ensembling model for semisupervised medical image segmentation},
  author={Li, Xiaomeng and Yu, Lequan and Chen, Hao and Fu, Chi-Wing and Xing, Lei and Heng, Pheng-Ann},
  journal={IEEE transactions on neural networks and learning systems},
  volume={32},
  number={2},
  pages={523--534},
  year={2020},
  publisher={IEEE}
}

@article{xu2023ambiguity,
  title={Ambiguity-selective consistency regularization for mean-teacher semi-supervised medical image segmentation},
  author={Xu, Zhe and Wang, Yixin and Lu, Donghuan and Luo, Xiangde and Yan, Jiangpeng and Zheng, Yefeng and Tong, Raymond Kai-yu},
  journal={Medical Image Analysis},
  volume={88},
  pages={102880},
  year={2023},
  publisher={Elsevier}
}

@ARTICLE{wu2023exploring,
  author={Wu, Huimin and Li, Xiaomeng and Cheng, Kwang-Ting},
  journal={IEEE Transactions on Neural Networks and Learning Systems}, 
  title={Exploring Feature Representation Learning for Semi-Supervised Medical Image Segmentation}, 
  year={2024},
  volume={35},
  number={11},
  pages={16589-16601},
  doi={10.1109/TNNLS.2023.3296652}}

@inproceedings{wang2023mcf,
  title={Mcf: Mutual correction framework for semi-supervised medical image segmentation},
  author={Wang, Yongchao and Xiao, Bin and Bi, Xiuli and Li, Weisheng and Gao, Xinbo},
  booktitle={Proceedings of the IEEE/CVF conference on computer vision and pattern recognition},
  pages={15651--15660},
  year={2023}
}

@inproceedings{miao2023caussl,
  title={Caussl: Causality-inspired semi-supervised learning for medical image segmentation},
  author={Miao, Juzheng and Chen, Cheng and Liu, Furui and Wei, Hao and Heng, Pheng-Ann},
  booktitle={Proceedings of the IEEE/CVF International Conference on Computer Vision},
  pages={21426--21437},
  year={2023}
}

@article{chen2022semi,
  title={Semi-supervised unpaired medical image segmentation through task-affinity consistency},
  author={Chen, Jingkun and Zhang, Jianguo and Debattista, Kurt and Han, Jungong},
  journal={IEEE Transactions on Medical Imaging},
  volume={42},
  number={3},
  pages={594--605},
  year={2022},
  publisher={IEEE}
}

@article{luo2022semi,
  title={Semi-supervised medical image segmentation via uncertainty rectified pyramid consistency},
  author={Luo, Xiangde and Wang, Guotai and Liao, Wenjun and Chen, Jieneng and Song, Tao and Chen, Yinan and Zhang, Shichuan and Metaxas, Dimitris N and Zhang, Shaoting},
  journal={Medical Image Analysis},
  volume={80},
  pages={102517},
  year={2022},
  publisher={Elsevier}
}

@article{wang2024boundary,
  title={Boundary-Aware Prototype in Semi-Supervised Medical Image Segmentation},
  author={Wang, YongChao and Xiao, Bin and Bi, Xiuli and Li, Weisheng and Gao, Xinbo},
  journal={IEEE Transactions on Image Processing},
  year={2024},
  publisher={IEEE}
}

@ARTICLE{10177922,
  author={Li, Jun and Zhu, Hongzhang and Chen, Tao and Qian, Xiaohua},
  journal={IEEE Journal of Biomedical and Health Informatics}, 
  title={Generalizable Pancreas Segmentation via a Dual Self-Supervised Learning Framework}, 
  year={2023},
  volume={27},
  number={10},
  pages={4780-4791},
  doi={10.1109/JBHI.2023.3294278}
}

@article{wang2021annotation,
  title={Annotation-efficient deep learning for automatic medical image segmentation},
  author={Wang, Shanshan and Li, Cheng and Wang, Rongpin and Liu, Zaiyi and Wang, Meiyun and Tan, Hongna and Wu, Yaping and Liu, Xinfeng and Sun, Hui and Yang, Rui and others},
  journal={Nature communications},
  volume={12},
  number={1},
  pages={5915},
  year={2021},
  publisher={Nature Publishing Group UK London}
}

@inproceedings{long2015fully,
  title={Fully convolutional networks for semantic segmentation},
  author={Long, Jonathan and Shelhamer, Evan and Darrell, Trevor},
  booktitle={Proceedings of the IEEE conference on computer vision and pattern recognition},
  pages={3431--3440},
  year={2015}
}

@INPROCEEDINGS{9761681,
  author={Thompson, Bethany H. and Di Caterina, Gaetano and Voisey, Jeremy P.},
  booktitle={2022 IEEE 19th International Symposium on Biomedical Imaging (ISBI)}, 
  title={Pseudo-Label Refinement Using Superpixels for Semi-Supervised Brain Tumour Segmentation}, 
  year={2022},
  volume={},
  number={},
  pages={1-5},
  doi={10.1109/ISBI52829.2022.9761681}}

@ARTICLE{9757875,
  author={Han, Kai and Liu, Lu and Song, Yuqing and Liu, Yi and Qiu, Chengjian and Tang, Yangyang and Teng, Qiaoying and Liu, Zhe},
  journal={IEEE Journal of Biomedical and Health Informatics}, 
  title={An Effective Semi-Supervised Approach for Liver CT Image Segmentation}, 
  year={2022},
  volume={26},
  number={8},
  pages={3999-4007},
  doi={10.1109/JBHI.2022.3167384}}

@article{qiu2023federated,
  title={Federated semi-supervised learning for medical image segmentation via pseudo-label denoising},
  author={Qiu, Liang and Cheng, Jierong and Gao, Huxin and Xiong, Wei and Ren, Hongliang},
  journal={IEEE journal of biomedical and health informatics},
  volume={27},
  number={10},
  pages={4672--4683},
  year={2023},
  publisher={IEEE}
}

@article{qiao2022semi,
  title={Semi-supervised CT lesion segmentation using uncertainty-based data pairing and SwapMix},
  author={Qiao, Pengchong and Li, Han and Song, Guoli and Han, Hu and Gao, Zhiqiang and Tian, Yonghong and Liang, Yongsheng and Li, Xi and Zhou, S Kevin and Chen, Jie},
  journal={IEEE Transactions on Medical Imaging},
  volume={42},
  number={5},
  pages={1546--1562},
  year={2022},
  publisher={IEEE}
}

@article{jing2020self,
  title={Self-supervised visual feature learning with deep neural networks: A survey},
  author={Jing, Longlong and Tian, Yingli},
  journal={IEEE transactions on pattern analysis and machine intelligence},
  volume={43},
  number={11},
  pages={4037--4058},
  year={2020},
  publisher={IEEE}
}

@inproceedings{chen2023decoupled,
  title={Decoupled consistency for semi-supervised medical image segmentation},
  author={Chen, Faquan and Fei, Jingjing and Chen, Yaqi and Huang, Chenxi},
  booktitle={International conference on medical image computing and computer-assisted intervention},
  pages={551--561},
  year={2023},
  organization={Springer}
}

@article{XIONG2021101832,
title = {A global benchmark of algorithms for segmenting the left atrium from late gadolinium-enhanced cardiac magnetic resonance imaging},
journal = {Medical Image Analysis},
volume = {67},
pages = {101832},
year = {2021},
doi = {https://doi.org/10.1016/j.media.2020.101832},
author = {Zhaohan Xiong and Qing Xia and Zhiqiang Hu and Ning Huang and Cheng Bian and Yefeng Zheng and Sulaiman Vesal and Nishant Ravikumar and Andreas Maier and Xin Yang and Pheng-Ann Heng and Dong Ni and Caizi Li and Qianqian Tong and Weixin Si and Elodie Puybareau and Younes Khoudli and Thierry Géraud and Chen Chen and Wenjia Bai and Daniel Rueckert and Lingchao Xu and Xiahai Zhuang and Xinzhe Luo and Shuman Jia and Maxime Sermesant and Yashu Liu and Kuanquan Wang and Davide Borra and Alessandro Masci and Cristiana Corsi and Coen {de Vente} and Mitko Veta and Rashed Karim and Chandrakanth Jayachandran Preetha and Sandy Engelhardt and Menyun Qiao and Yuanyuan Wang and Qian Tao and Marta Nuñez-Garcia and Oscar Camara and Nicolo Savioli and Pablo Lamata and Jichao Zhao}
}
\bibliographystyle{icml2026}

\newpage
\appendix
\onecolumn
\section{Data Preprocessing.}
\label{appA}
The LA benchmark contains 100 gadolinium-enhanced cardiac MRI volumes, acquired with an isotropic spacing of (0.625 × 0.625 × 0.625) mm. The Pancreas-CT dataset provides 82 contrast-enhanced abdominal CT scans with a fixed in-plane size of (512 × 512) and slice thickness between 1.5 and 2.5 mm. BraTS2019 includes multi-parametric brain MRI from 335 glioma subjects, where each case offers four sequences together with voxel-wise tumor annotations. For BraTS, we follow the protocol adopted by recent semi-supervised studies and report whole-tumor (WT) segmentation using the FLAIR modality; we use 250/25/60 volumes for training/validation/testing. To reduce the influence of anisotropic spacing on 3D convolutions, all scans are resampled to an isotropic resolution of (1.0 × 1.0 × 1.0) mm, and we keep the same preprocessing pipeline and data partitioning (including labeled/unlabeled splits) as the canonical settings in prior works.

\section{Implementation Details.}
\label{appB}
All experiments are conducted in PyTorch on the NVIDIA 4080 GPU. The framework is trained with an SGD optimizer for 20,000 iterations at a learning rate of 0.01, and a batch size of four—comprising two labeled and two unlabeled images. To avoid memory overflow, we adopt a sliding window approach rather than feeding the entire 3D volume directly. Specifically, sub-volumes of 112 × 112 × 80 (LA cases, the strides are set to 18 × 18 × 4) and 96 × 96 × 96 (Pancreas-NIH and BraTS2019 cases, the strides are set to 16 × 16 × 16) were randomly cropped as training inputs. During testing, windows of the same size were employed to perform sliding inference across the 3D volumes, and the results from each window were subsequently aggregated to obtain the final segmentation. 

\section{Evaluation Metrics}
\label{appC}
We report three commonly used metrics: the Dice similarity coefficient (DSC), the 95th-percentile Hausdorff distance (95HD), and the average surface distance (ASD). DSC measures volumetric agreement between the prediction and the reference by comparing their overlap. ASD summarizes the mean discrepancy between the two surfaces, while 95HD reflects near-worst-case boundary errors, which reduces sensitivity to a few extreme outliers compared with the maximum Hausdorff distance. Better segmentation corresponds to higher DSC and lower ASD/95HD.

\section{Detailed Results of SOTA Methods}
\label{appD}

\begin{table*}[h]
\caption{Comparisons with other semi-supervised methods on BraTS2019 dataset.}
\centering
\small
\begin{threeparttable}
\begin{tabular}{lccccc}
\toprule
\multirow{2.4}{*}{Method} & \multirow{2.4}{*}{Protocol} & Scans & \multicolumn{3}{c}{Metrics} \\
\cmidrule{3-3} \cmidrule{4-6}
 &  & Labeled / Unlabeled & DSC (\%) $\uparrow$  & ASD (mm) $\downarrow$ & 95HD (mm) $\downarrow$ \\
\midrule
UG-MCL \cite{zhang2023uncertainty} & last / best & 25 / 225 & 82.82 & 2.30 & 11.29 \\
CPCL \cite{9741294}               & last / best & 25 / 225 & 83.36 & 1.99 & 11.74 \\
CauSSL \cite{miao2023caussl}      & last / best & 25 / 225 & 83.54 & 1.98 & 12.53 \\
AC-MT \cite{xu2023ambiguity}      & last / best & 25 / 225 & 83.77 & 1.93 & 11.37 \\
URPC \cite{luo2022semi}           & last / best & 25 / 225 & 84.16 & 2.63 & 11.01 \\
MRPL \cite{su2024mutual}          & last / best & 25 / 225 & 84.29 & 2.55 & 9.57  \\
DBiSL \cite{LI2026131439}         & last / best & 25 / 225 & 85.09 & 1.89 & 8.12  \\
BoANet \cite{10422862}            & last / best & 25 / 225 & 85.43 & 2.15 & 8.69  \\
TraCoCo \cite{liu2024translation} & last / best & 25 / 225 & \underline{85.71} & 2.27 & 9.20  \\
Our median                        & last / best & 25 / 225 & 85.27 & 3.02 & 11.67 \\
Our maximum                       & last / best & 25 / 225 & \textbf{86.52} & 3.59 & 11.48 \\
\midrule
UG-MCL \cite{zhang2023uncertainty} & last / best & 50 / 200 & 83.61 & 2.26 & 11.44 \\
AC-MT \cite{xu2023ambiguity}       & last / best & 50 / 200 & 84.63 & 2.11 & 9.50  \\
MRPL \cite{su2024mutual}           & last / best & 50 / 200 & 85.47 & 2.00 & 7.76  \\
URPC \cite{luo2022semi}            & last / best & 50 / 200 & 85.49 & 2.04 & 8.47  \\
BoANet \cite{10422862}             & last / best & 50 / 200 & 85.88 & 1.92 & 8.26  \\
SFR \cite{10847777}                & last / best & 50 / 200 & 86.09 & 2.47 & 8.51  \\
CDS \cite{hu2025competitive}                & last / best & 50 / 200 & 86.12 & 1.83 & 7.78  \\
TraCoCo \cite{liu2024translation}  & last / best & 50 / 200 & \textbf{86.69} & 1.93 & 8.04  \\
Our median                         & last / best & 50 / 200 & 86.47 & 2.44 & 9.30  \\
Our maximum                        & last / best & 50 / 200 & \underline{86.68} & 2.00 & 7.11  \\
\bottomrule
\end{tabular}

\begin{tablenotes}[flushleft]
\footnotesize
\item \textbf{Note:} BraTS2019 provides a validation set, thus the results under the two protocols coincide.
\end{tablenotes}
\end{threeparttable}

\vspace{-2pt}
\label{tab:taba1}
\vspace{-6pt}
\end{table*}

\begin{table*}[ht]
\caption{Comparisons with other semi-supervised methods on LA dataset.}
\centering
\small
\begin{tabular}{lcccccc}
\toprule
\multirow{2.4}{*}{Method} & \multirow{2.4}{*}{Protocol} & \multicolumn{1}{c}{Scans} & \multicolumn{3}{c}{Metrics} \\
\cmidrule{4-6}
  & & Labeled / Unlabeled & DSC (\%) $\uparrow$ & ASD (mm) $\downarrow$ & 95HD (mm) $\downarrow$ \\
\midrule

TAC \cite{chen2022semi} & last & 8 / 72 & 84.73 & 2.72 & 11.45 \\
CDS \cite{hu2025competitive} & last & 8 / 72 & 87.21 & 2.97 & 11.02  \\
DTC \cite{luo2021semi} & last & 8 / 72 & 87.51 & 2.36 & 8.23 \\
DBiSL \cite{LI2026131439} & last & 8 / 72 & 88.78 & / & / \\
TraCoCo \cite{liu2024translation} & last & 8 / 72 & 89.29 & 2.28 & 6.92 \\
ARCO-SG \cite{you2023rethinking} & last & 8 / 72 & 89.90 & 1.47 & / \\
Our median & last & 8 / 72 & \underline{90.28} & 1.77 & 6.49 \\
Our maximum & last & 8 / 72 & \textbf{90.39} & 1.88 & 6.40 \\
\midrule

SGTC \cite{yan2025sgtc} & best & 8 / 72 & 84.70 & 4.26 & 17.44 \\
AUSS \cite{adiga2024anatomically} & best & 8 / 72 & 86.58 & / & 11.82 \\
CPL \cite{LV2026112321} & best & 8 / 72 & 87.01 & 2.97 & 9.05 \\
CRDS \cite{aralikatti2023dual} & / & 8 / 72 & 88.51 & 2.13 & 6.86 \\
BOA-Net \cite{10422862} & / & 8 / 72 & 88.91 & 2.11 & 7.73 \\
MC-Net+ \cite{wu2022mutual} & best & 8 / 72 & 88.96 & 1.86 & 7.93 \\
VerSemi \cite{10945994} & best & 8 / 72 & 89.01 & 2.57 & 9.03 \\
AC-MT \cite{xu2023ambiguity} & best & 8 / 72 & 89.12 & 2.19 & 11.05 \\
Co-BioNet \cite{peiris2023uncertainty} & best & 8 / 72 & 89.20 & 1.90 & 6.44 \\
ICSSL \cite{10530285} & best & 8 / 72 & 89.25 & 1.86 & 6.96 \\
BCP \cite{bai2023bidirectional} & best & 8 / 72 & 89.62 & 1.76 & 6.81 \\
TraCoCo \cite{liu2024translation} & best & 8 / 72 & 89.86 & 2.01 & 6.81 \\
MRPL \cite{su2024mutual} & best & 8 / 72 & 89.86 & 1.85 & / \\
w2SPC \cite{yang2025semi} & / & 8 / 72  & 90.23 & 1.95 & 7.16  \\
OTCMC \cite{guo2025optimal} & / & 8 / 72 & 90.26 & 1.65 & 5.96 \\
A\&D \cite{wang2023towards} & best & 8 / 72 & 90.31 & 1.64 & 5.55 \\
CC-Net \cite{wang2025data} & best & 8 / 72 & \underline{90.82} & 1.72 & 7.50 \\
Our median & best & 8 / 72 & 90.73 & 1.77 & 5.87 \\
Our maximum & best & 8 / 72 & \textbf{90.85} & 1.80 & 5.45 \\
\midrule

TAC \cite{chen2022semi} & last & 16 / 64 & 87.75 & 2.04 & 9.45 \\
MCF \cite{wang2023mcf} & last & 16 / 64 & 88.71 & 1.90 & 6.32 \\
DTC \cite{luo2021semi} & last & 16 / 64 & 89.42 & 2.10 & 7.32 \\
CDS \cite{hu2025competitive} & last & 16 / 64 & 90.33 & 1.91 & 6.16  \\
BSNet \cite{10375506} & last & 16 / 64 & 90.43 & 1.63 & 6.21 \\
DBiSL \cite{LI2026131439}  & last & 16 / 64 & 90.54 & 1.80 & 6.05 \\
TraCoCo \cite{liu2024translation} & last & 16 / 64 & 90.94 & 1.79 & 5.49 \\
SFR \cite{10847777} & last & 16 / 64 & 91.00 & 1.61 & 6.13 \\
AUA \cite{wu2023exploring} & last & 16 / 64 & \underline{91.08} & 1.80 & 5.60 \\
Our median & last & 16 / 64 & 90.83 & 1.63 & 6.03 \\
Our maximum & last & 16 / 64 & \textbf{91.36} & 1.86 & 6.16 \\
\midrule

AUSS \cite{adiga2024anatomically} & best & 16 / 64 & 88.60 & / & 7.61 \\
BaPC \cite{wang2024boundary} & best & 16 / 64 & 89.71 & 1.85 & 6.08 \\
UG-MCL \cite{zhang2023uncertainty} & best & 16 / 64 & 90.16 & 1.98 & 6.50 \\
AC-MT \cite{xu2023ambiguity} & best & 16 / 64 & 90.31 & 1.76 & 6.21 \\
OTCMC \cite{guo2025optimal} & / & 16 / 64 & 90.86 & 1.60 & 6.27 \\
VerSemi \cite{10945994} & best & 16 / 64 & 90.89 & 1.72 & 5.38 \\
MRPL \cite{su2024mutual} & best & 16 / 64 & 91.02 & 1.66 & 5.78 \\
w2SPC \cite{yang2025semi}& / & 16 / 64 & 91.05 & 1.67 & 5.94 \\
MC-Net+ \cite{wu2022mutual} & best & 16 / 64 & 91.07 & 1.67 & 5.84 \\
Co-BioNet \cite{peiris2023uncertainty} & best & 16 / 64 & 91.26 & 1.64 & 5.17 \\
BoA-Net \cite{10422862} & / & 16 / 64 & 91.38 & 1.54 & 5.09 \\
TraCoCo \cite{liu2024translation} & best & 16 / 64 & 91.51 & 1.79 & 5.63 \\
CC-Net \cite{wang2025data} & best & 16 / 64 & \underline{91.60} & 1.54 & 5.27 \\
Our median & best & 16 / 64 & 91.44 & 1.64 & 5.42 \\
Our maximum & best & 16 / 64 & \textbf{91.62} & 1.62 & 5.44 \\
\bottomrule
\end{tabular}
\label{tab:taba2}
\end{table*}

\begin{table*}[t] 
\caption{Comparisons with other semi-supervised methods on Pancreas-NIH dataset.}
\centering
\small
\begin{tabular}{lccccc}
\toprule
\multirow{2.4}{*}{Method} & \multicolumn{1}{c}{Protocol} & \multicolumn{1}{c}{Scans used} & \multicolumn{3}{c}{Metrics} \\
\cmidrule{2-3} \cmidrule{4-6}
 &  & Labeled / Unlabeled & DSC (\%) $\uparrow$  & ASD (mm) $\downarrow$ & 95HD (mm) $\downarrow$ \\
\midrule

CauSSL \cite{miao2023caussl}        & last & 6 / 56 & 72.89 & 4.37 & 14.19 \\
TraCoCo \cite{liu2024translation}  & last & 6 / 56 & 79.22 & 2.57 & 8.46 \\
Our median                         & last & 6 / 56 & \underline{81.08} & 3.55 & 9.33 \\
Our maximum                        & last & 6 / 56 & \textbf{81.20} & 4.48 & 16.11 \\
\midrule

BaPC \cite{wang2024boundary}        & best & 6 / 56 & 68.64 & 3.59 & 16.37 \\
SC-SSL \cite{10328616}             & best & 6 / 56 & 70.09 & 1.97 & 12.88 \\
MC-Net+ \cite{wu2022mutual}        & best & 6 / 56 & 74.01 & 3.34 & 12.59 \\
URPC \cite{luo2022semi}            & best & 6 / 56 & 74.89 & 3.74 & 11.30 \\
MRPL \cite{su2024mutual}           & best & 6 / 56 & 75.93 & 1.54 & 9.07 \\
BoA-Net \cite{10422862}            & / & 6 / 56 & 76.06 & 3.62 & 11.23 \\
Co-BioNet \cite{peiris2023uncertainty} & best & 6 / 56 & 77.89 & 1.39 & 8.81 \\
VerSemi \cite{10945994}            & best & 6 / 56 & 78.62 & 2.28 & 7.99 \\
CC-Net \cite{wang2025data}         & best & 6 / 56 & 81.93 & 1.98 & 7.79 \\
Our median                         & best & 6 / 56 & \underline{82.22} & 2.50 & 8.16 \\
Our maximum                        & best & 6 / 56 & \textbf{82.52} & 1.89 & 5.50 \\
\midrule

MCF \cite{wang2023mcf}             & last & 12 / 50 & 75.00 & 3.27 & 11.59 \\
DTC \cite{luo2021semi}             & last & 12 / 50 & 78.27 & 2.25 & 8.36 \\
AUA \cite{wu2023exploring}         & last & 12 / 48 & 79.81 & 1.64 & 5.90 \\
CauSSL \cite{miao2023caussl}       & last & 12 / 50 & 80.92 & 1.53 & 8.11 \\
DBiSL \cite{LI2026131439}          & last & 12 / 50 & 81.09 & 1.71 & 6.90 \\
TraCoCo \cite{liu2024translation}  & last & 12 / 50 & 81.80 & 1.49 & 5.70 \\
Our median                         & last & 12 / 50 & \underline{83.44} & 1.61 & 4.90 \\
Our maximum                        & last & 12 / 50 & \textbf{84.14} & 2.14 & 5.62 \\
\midrule

BaPC \cite{wang2024boundary}       & best & 12 / 50 & 76.36 & 2.83 & 10.59 \\
CoraNet \cite{shi2021inconsistency} & best & 12 / 50 & 79.67 & 1.89 & 7.59 \\
URPC \cite{luo2022semi}            & best & 12 / 50 & 80.31 & 2.10 & 6.58 \\
MC-Net+ \cite{wu2022mutual}        & best & 12 / 50 & 80.59 & 1.74 & 6.47 \\
SC-SSL \cite{10328616}             & best & 12 / 50 & 80.76 & 1.73 & 6.79 \\
BoA-Net \cite{10422862}             & / & 12 / 50 & 80.80 & 1.65 & 6.01 \\
ICSSL \cite{10530285}              & best & 12 / 50 & 81.17 & 1.46 & 6.17 \\
MRPL \cite{su2024mutual}           & best & 12 / 50 & 81.53 & 1.33 & 6.81 \\
Co-BioNet \cite{peiris2023uncertainty} & best & 12 / 50 & 82.22 & 1.16 & 7.71 \\
BCP \cite{bai2023bidirectional}    & best & 12 / 50 & 82.91 & 2.25 & 6.43 \\
TraCoCo \cite{liu2024translation}  & best & 12 / 50 & 83.36 & 1.74 & 7.34 \\
VerSemi \cite{10945994}            & best & 12 / 50 & 83.55 & 1.35 & 5.02 \\
CC-Net \cite{wang2025data}         & best & 12 / 50 & 83.89 & 1.81 & 5.24 \\
Our median                         & best & 12 / 50 & \underline{83.94} & 2.26 & 6.25 \\
Our maximum                        & best & 12 / 50 & \textbf{84.14} & 2.14 & 5.62 \\

\bottomrule
\end{tabular}
\vspace{-2pt}
\label{tab:taba3}
\vspace{-6pt}
\end{table*}


\end{document}